\documentclass[conference,a4paper]{IEEEtran}

\usepackage{color}
\ifCLASSINFOpdf
   \usepackage[pdftex]{graphicx}
\else
\fi
\usepackage{url}
\usepackage{graphicx}
\usepackage{amsmath,amssymb} % define this before the line numbering.
\usepackage{color}
\usepackage[utf8]{inputenc}
\usepackage[english]{babel}
\usepackage{subcaption}
\usepackage[capitalise]{cleveref}
\usepackage{booktabs}
\usepackage{tabularx}
\usepackage[table]{xcolor}

\begin{document}

\newcolumntype{Y}{>{\centering\arraybackslash}X}

\newsavebox{\dualsegnet}

% *** Do not adjust lengths that control margins, column widths, etc. ***
% *** Do not use packages that alter fonts (such as pslatex).         ***
% There should be no need to do such things with IEEEtran.cls V1.6 and later.
% (Unless specifically asked to do so by the journal or conference you plan
% to submit to, of course. )

% correct bad hyphenation here
\hyphenation{op-tical net-works semi-conduc-tor}
\IEEEoverridecommandlockouts

\title{Fusion of Heterogeneous Data in Convolutional Networks for Urban Semantic Labeling}

\author{\IEEEauthorblockN{Nicolas Audebert\IEEEauthorrefmark{1}\IEEEauthorrefmark{2}, Bertrand Le Saux\IEEEauthorrefmark{1}}
\IEEEauthorblockA{\IEEEauthorrefmark{1}ONERA, \textit{The French Aerospace Lab}\\
F-91761 Palaiseau, France \\
\texttt{\{nicolas.audebert,bertrand.le\_saux\}@onera.fr}}
\and
\IEEEauthorblockN{Sébastien Lefèvre\IEEEauthorrefmark{2}}
\IEEEauthorblockA{\IEEEauthorrefmark{2}Univ. Bretagne-Sud, UMR 6074, IRISA\\
F-56000 Vannes, France\\
\texttt{sebastien.lefevre@irisa.fr}}}


%Telephone: (800) 555--1212, Fax: (888) 555--1212}

\IEEEspecialpapernotice{(Invited Paper)}

\IEEEpubid{978-1-5090-5808-2/17/\$31.00~ \copyright~2017 IEEE}
%\pubid{978-1-5090-5808-2/17/\$31.00~ \copyright~2017 Crown}%For Crown government employees
%\pubid{978-1-5090-5808-2/17/\$31.00~ \copyright~2017 European Union}%For European Union employees
%\pubid{U.S. Government work not protected by U.S. copyright}%US government employees

\maketitle

\begin{abstract}
In this work, we present a novel module to perform fusion of heterogeneous data using fully convolutional networks for semantic labeling. We introduce residual correction as a way to learn how to fuse predictions coming out of a dual stream architecture. Especially, we perform fusion of DSM and IRRG optical data on the ISPRS Vaihingen dataset over a urban area and obtain new state-of-the-art results.
\end{abstract}


% \fi
\IEEEpeerreviewmaketitle

\section{Introduction}

Following the take over of deep larning over the computer vision field, deep convolutional neural networks (CNN) propagated to remote sensing image processing. Deep networks are now state-of-the-art for object detection and classification, but also for semantic labeling, both in everyday images, e.g. PASCAL VOC2012~\cite{everingham_pascal_2014}, and Earth Observation data, e.g. ISPRS Vaihingen 2D Semantic Labeling Challenge~\cite{rottensteiner_isprs_2012}. However, these deep networks have been originally designed for everyday RGB images. On the contrary, remote sensing data is rarely limited neither to RGB, nor to optical data, and often combines several heterogeneous sensors. In scene understanding of Earth Observation images, data fusion can therefore significantly improve a statistical model's accuracy by combining specific information from the different sensors. For example, hyperspectral and LiDAR sensors have different spatial resolution and do not share the same physical properties, although both the spectrum and the measured height can be relevant features for classification. In this work, we present a new residual correction module designed to perform efficient data fusion using CNN. We apply this technique to the IRRG images and DSM data of the ISPRS Vaihingen dataset and obtain new state-of-the-art results.

\section{Related Work}

Most works related to deep learning for urban semantic labeling use 3-channels networks designed for RGB (and sometimes IRRG), fine-tuned from a model trained on the ImageNet dataset~\cite{paisitkriangkrai_effective_2015,marmanis_semantic_2016,marmanis_deep_2016}. Dual stream neural networks for data fusion have been introduced in~\cite{ngiam_multimodal_2011} in an unsupervised framework for joint audio-video representation learning, using a dual stream auto-encoder. The same principles have been transposed to supervised learning in~\cite{eitel_multimodal_2015} for classification of RGB-D data.%, where the predictions of two parallel CNN are fused by fine-tuning a large fully connected layer that spans merges the two output vectors.

Data fusion using CNN for classification of remote sensing images has also been explored in the Data Fusion Contest (DFC) 2015, where CNN have been used for multimodal and multi-scale feature extraction in combination with a SVM classifier~\cite{lagrange_benchmarking_2015}. Semantic labeling on the ISPRS Vaihingen dataset was further improved using fusion of CNN-based and expert-crafted features with random forests~\cite{paisitkriangkrai_effective_2015}. In the DFC 2016, semantic labeling based on a high resolution multispectral image and tracklet analysis on a spaceborne video were combined for traffic density and activity analysis~\cite{mou_spatiotemporal_2016}.

Finally, residual learning~\cite{he_deep_2016} was introduced with the idea that deep networks have trouble learning the identity function. Using bypass connections, the network would only have to learn a residual addition to the input, which would be easier.

Building on these works, our residual correction is a generic module fully integrated in the CNN pipeline that can be added on any multiple stream architecture. Moreover, it uses recent advances in deep learning by linking residual learning to the signal processing viewpoint on error correction. Especially, we integrate the fusion with the recent fully convolutional networks (FCN)~\cite{long_fully_2015} that are able to perform end-to-end dense semantic labeling.

\section{Heterogeneous Data Fusion with Residual Correction}

\begin{figure*}[t]
  \centering
  \savebox{\dualsegnet}{\includegraphics[width=0.43\textwidth]{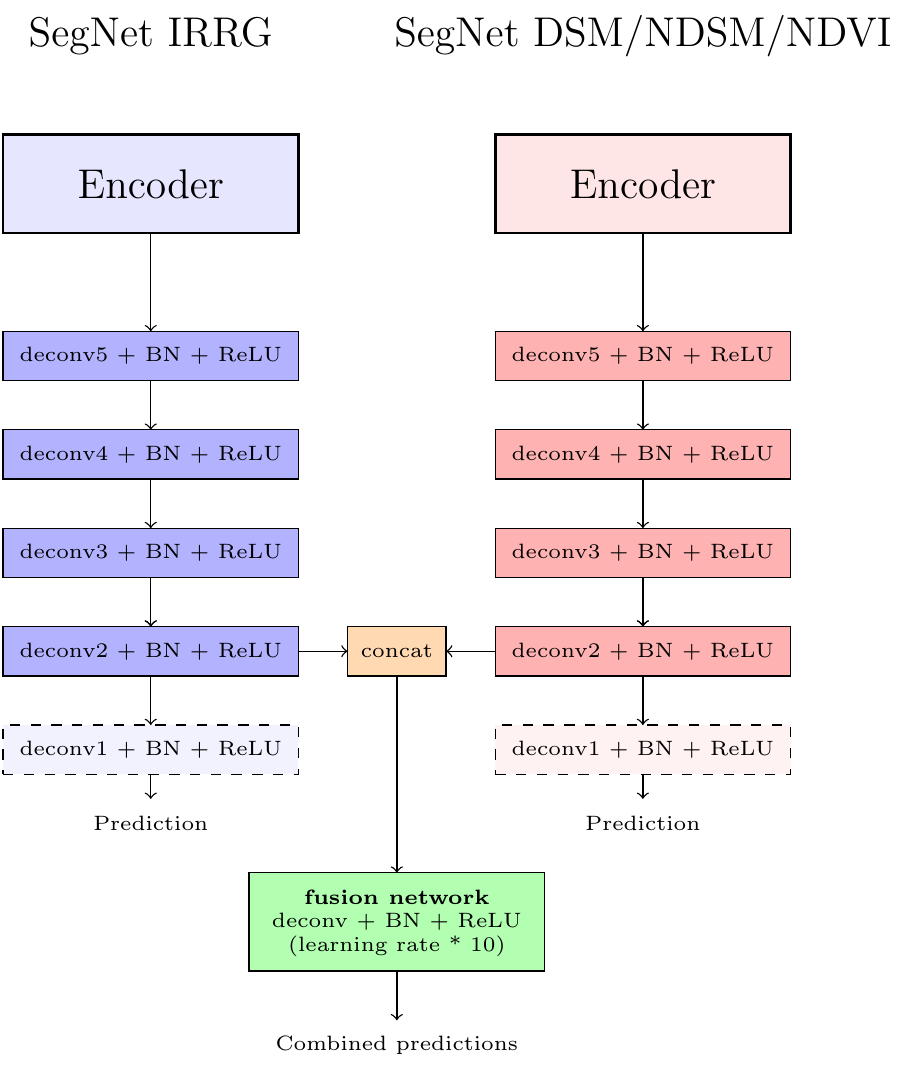}}
    \begin{subfigure}[b]{0.43\textwidth}
        \centering
        \raisebox{\dimexpr.5\ht\dualsegnet-0.42\height}{
          \includegraphics[width=\textwidth]{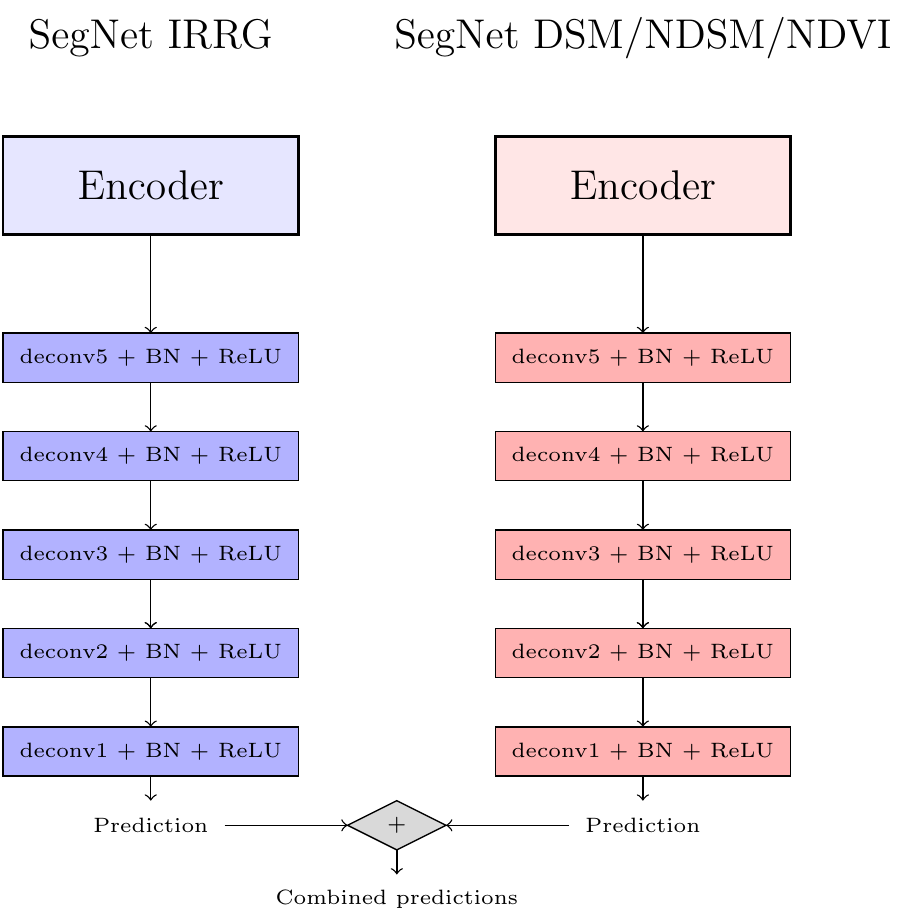}}
        \caption{Averaging strategy.}
        \label{fig:fusion_segnet_sum}
  \end{subfigure}
  \begin{subfigure}[b]{0.43\textwidth}
        \centering
        \usebox{\dualsegnet}
        \caption{Fusion network strategy.}
        \label{fig:fusion_segnet_network}
  \end{subfigure}
  \caption{Fusion strategies of our dual-stream SegNet architecture.}
  \label{fig:fusion_segnet}
\end{figure*}

\begin{figure}[t]
  \centering
  \includegraphics[height=0.49\textwidth, angle=90]{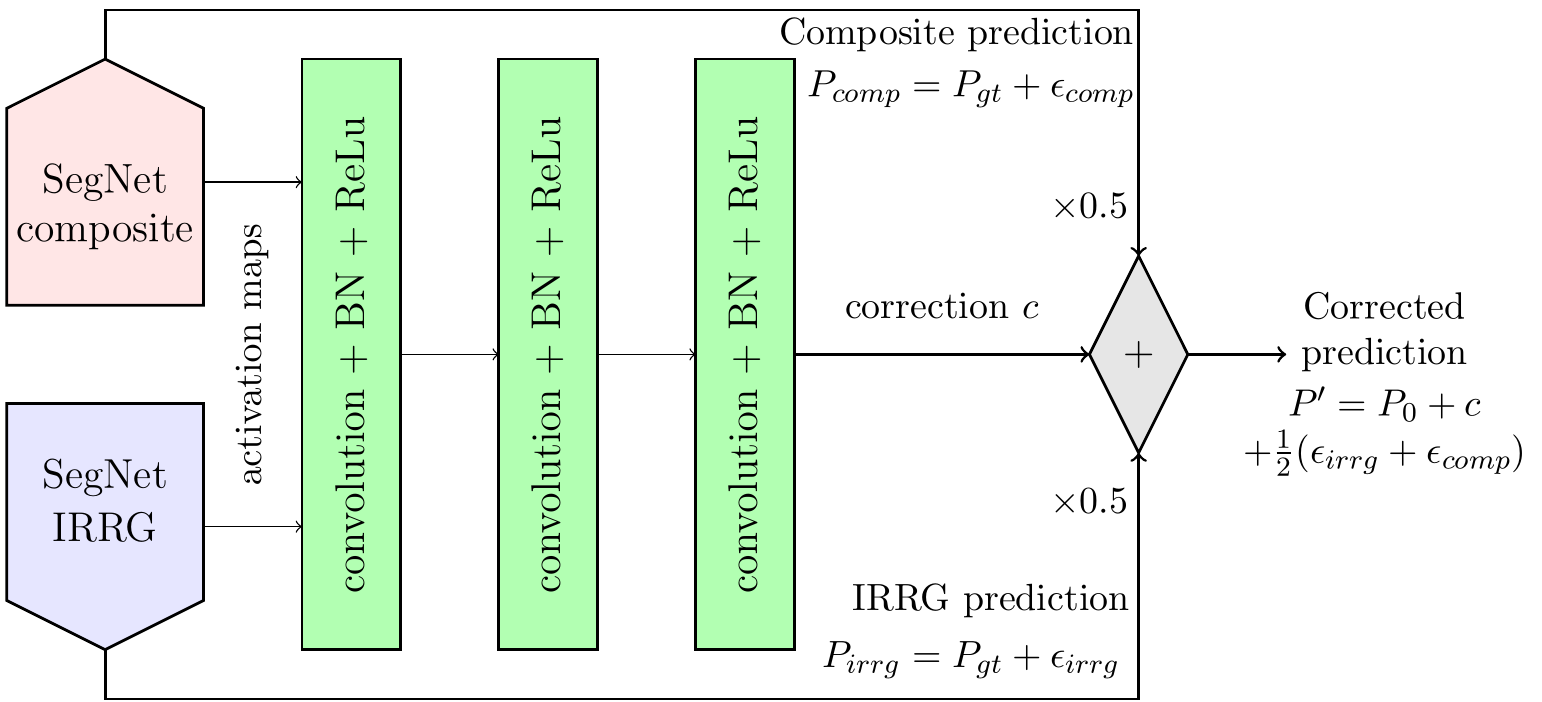}
  \caption{Fusion network to correct predictions with information from complementary SegNets using heterogeneous data.}
  \label{fig:correction_network}
\end{figure}

\IEEEpubidadjcol

A naive approach to data fusion using deep networks would be to concatenate all channels (e.g. RGB and depth) and use it as the input. However, preliminary experiments showed that this actually degrades the network accuracy as heterogeneous data sources require different processing. Luckily, prediction oriented fusion has been proven effective using dual stream networks both in unsupervised~\cite{ngiam_multimodal_2011} and supervised~\cite{eitel_multimodal_2015} settings. Therefore, we first train two 3-channels SegNet~\cite{badrinarayanan_segnet:_2015} for semantic labeling using our two data sources. Then, we perform prediction fusion at end of the networks by merging the two parallel streams. As a baseline, we just perform simple averaging after the softmax (\cref{fig:fusion_segnet_sum}). To improve the fusion, we introduce a fusion neural network that learns to improve the average prediction by using data specific information.

Building on the idea of residual deep learning~\cite{he_deep_2016}, we propose a fusion network based on residual correction. We define a convolutional residual block using the same parameters as the rest of the SegNet network ($3\times3$ convolutions and $1$ pixel of padding). Intermediate feature maps from the decoding parts of the two SegNets are fed as inputs to a 3-convolution layers ``correction'' network (cf. \cref{fig:fusion_segnet_network}. The output of the residual block is then summed in the residual fashion with the average of the two predictions, as illustrated by \cref{fig:fusion_segnet_network}. Residual learning fits this use case, as the average score is already a close estimate of the truth. To improve the results, we aim to use the complementary channels to correct small errors in the prediction maps. In this context, the residual can be seen as a corrective term for our predictive model. This module is trained using backpropagation on the standard softmax loss. Learning rates for the input SegNets are set to zero, as this considerably speeds up the training without significant loss.

%Let $M_r$ denote the input of the $r^{th}$ stream ($r \in \{1,\dots,R\}$ with $R = 2$ here), $P_{r}$ the output probability tensor and $Z_r$ the intermediate feature map used for the correction. The corrected prediction is:

% \begin{equation}
% P'(M_1, \dots, M_R) = P(M_1, \dots, M_R) + correction(Z_1, \dots, Z_R)
% \end{equation}
% where
% \begin{equation}
% P(M_1, \dots, M_n) = \frac{1}{R}\sum_{r=1}^R P_r(M_r)~.
% \end{equation}

Assuming that $P_0$ is the ground truth tensor and $P_i$ is the predicted output of the $i^{th}$ stream, we have :
\begin{equation}
P_i = P_0 + \epsilon_i \text{ where } \lvert \epsilon_i \lvert  \ll \rvert P_i \rvert
\end{equation}

$\epsilon_i$ is an error term that is small if the prediction $P_i$ is accurate enough. We expect the network to learn to estimate the errors and to infer when and how to merge the streams.

Let $R$ be the number of outputs on which to perform residual correction. We predict $P'$, the sum of the averaged predictions and the correction term $c$:
\begin{equation}
P' = P_{avg} + c = \frac{1}{R} \sum_{i=1}^R P_i + c = P_0 + \frac{1}{R} \sum_{i=1}^R \epsilon_i + c
\end{equation}

As our residual correction module is optimized to minimize the loss, we enforce:
\begin{equation}
\lVert P' - P_0 \rVert \rightarrow 0
\end{equation}
which translates into a constraint on $c$ and $\epsilon_i$:
\begin{equation}
\lVert \frac{1}{R} \sum_{i=1}^R \epsilon_i - c \rVert \rightarrow 0
\end{equation}

This can be seen as learning a model of the average error based on the feature maps. Indeed, at training time, the ground truth $P_0$ is known and the residual correction learns how to infer $\sum_{i=1}^R \epsilon_i$. The residual learning framework suits well this idea of error correction, as the residual is expected to be of a small amplitude compared to the main identity (or ``bypass'') signal. The residual correction module is detailed in \cref{fig:correction_network}.

\section{Experiments}

\subsection{Experimental Setup}

We test our method on the ISPRS Vaihingen 2D Semantic Labeling dataset comprised of IRRG images over an urban area. The 3 channels (i.e. near-infrared, red and green) are processed as an RGB image in the first stream of our dual SegNet architecture. The dataset also includes a Digital Surface Model (DSM) acquired with a Lidar and the Normalized Digital Surface Model (NDSM) from~\cite{gerke_use_2015}. We compute the Normalized Difference Vegetation Index (NDVI) from the near-infrared and red channels, which is an indicator for vegetation ($NDVI = {(IR - R)/(IR + R)}$).

For each tile, we aggregate DSM, NDSM and NDVI into a composite image used in the second stream of our architecture. The two streams use mostly heterogeneous data (height versus optical data). The composite image also includes redundant optical data (the NDVI) so that it contains relevant for information for all the classes (e.g. height helps discriminate road vs building while NDVI helps find the vegetation). Therefore, both the IRRG and composite images can be used to infer segmentations with similar accuracies, which will ease the predictions fusion.

We process the tiles using a $128 \times 128$ sliding window with a $32$px stride. We split the tiles with the public ground truth into a training set (12 tiles) and a validation set (4 tiles). We train separately the two SegNets for 10 epochs with a learning rate of 0.1, divided by 10 after 5 epochs. For our baseline, we compute the average prediction during testing. We fine-tune the residual correction module for 1 epoch, as longer training does not improve convergence.

\subsection{Results}

% \begin{table}[t]
%   \centering
%   \caption{Results on the validation set with different initialization policies.}
%   \begin{tabularx}{\textwidth}{ c | Y | Y | Y | Y | Y }
%   \toprule
%   Initialization & Random & \multicolumn{4}{c}{VGG-16}\\
%   \midrule
%   Learning rate ratio $\frac{lr_{enc}}{lr_{dec}}$ & 1 & 1 & 0.5 & 0.1 & 0\\
%   \midrule
%   Accuracy & 87.0\% & 87.2\% & \textbf{87.8\%} & 86.9\% & 86.5\%\\
%   \bottomrule
%   \end{tabularx}
%   \label{tab:initialization_results}
% \end{table}

\begin{table}[t]
  \centering
  \caption{Results on the validation set.}
  \setlength{\tabcolsep}{8pt}
  \begin{tabularx}{0.49\textwidth}{ Y c c c }
  \toprule
  %Type/Stride (px) & 128 {\small (no overlap)} & 64 {\small (50\% overlap)} & 32 {\small (75\% overlap)}\\
  Type/Stride (px) & 128 & 64 & 32\\
  \midrule
  Single stream (IRRG) & 87.8\% & 88.3\% & 88.8\%\\
  %Multi-kernel & 88.2\% & 88.6\% & 89.1\%\\
  Fusion (average) & 88.2\% & 88.7\% & 89.1\%\\
  Fusion (correction) & 88.6\% & 89.0\% & 89.5\%\\
  %Multi-kernel + Average & 88.5\% & 89.0\% & 89.5\%\\
  %Multi-kernel + Correction & 88.7\% & 89.3\% & \textbf{89.8\%}\\
  \bottomrule
  \end{tabularx}
  \label{tab:validation_results}
\end{table}

\begin{table}[t]
  \centering
  \caption{ISPRS 2D Semantic Labeling Vaihingen results.}
  \setlength\tabcolsep{1.5pt}
  \rowcolors{2}{gray!15}{white}
  \begin{tabularx}{0.49\textwidth}{ Y c c c c c c }
  \toprule
  Method & imp surf & building & low veg & tree & car & OA \\
  \midrule
%  Stair Vision Library {\scriptsize (``SVL\_3'')}\cite{gerke_use_2015} & 86.6\% &	91.0\% &	77.0\% &	85.0\%	& 55.6\% &	84.8\% \\
  RF + CRF {\scriptsize (``HUST'')}\cite{quang_efficient_2015} & 86.9\% & 92.0\% &	78.3\% &	86.9\% &	29.0\% &	85.9\% \\
%  CNN ensemble {\scriptsize (``ONE\_5'')}\cite{boulch_dag_2015} & 87.8\% &	92.0\% &	77.8\% &	86.2\% &	50.7\% &	85.9\% \\
%  FCN {\scriptsize (``UZ\_1'')} & 89.2\% &	92.5\% &	81.6\% &	86.9\% &	57.3\% &	87.3\% \\
%  FCN {\scriptsize (``UOA'')}\cite{lin_efficient_2015} & 89.8\% &	92.1\% &	80.4\% &	88.2\% &	82.0\% &	87.6\% \\
  CNN+RF+CRF {\tiny (``ADL\_3'')}\cite{paisitkriangkrai_effective_2015} & 89.5\% &	93.2\% &	82.3\% &	88.2\% &	63.3\% &	88.0\% \\
  FCN {\scriptsize (``DLR\_2'')}\cite{marmanis_semantic_2016} & 90.3\% &	92.3\% &	82.5\% &	89.5\% &	76.3\% &	88.5\% \\
  FCN+RF+CRF {\scriptsize (``DST\_2'')} & 90.5\% &	93.7\% &	83.4\% &	89.2\% &	72.6\% &	89.1\% \\
  \midrule
  Ours (IRRG only) & \textbf{91.5}\% &	94.3\% &	82.7\% &	89.3\% &	\textbf{85.7}\% &	89.4\% \\
  Ours (fusion) & 91.0\% &	\textbf{94.5}\% &	\textbf{84.4}\% &	\textbf{89.9}\% &	77.8\% &	\textbf{89.8}\% \\
  \bottomrule
  \end{tabularx}
  \label{tab:leaderboard}
\end{table}

\begin{figure}[t]
  \centering
  \captionsetup[subfigure]{singlelinecheck=off,justification=centering}
  \captionsetup[subfigure]{labelformat=empty}
  \begin{subfigure}[t]{0.15\textwidth}
    \includegraphics[width=\textwidth]{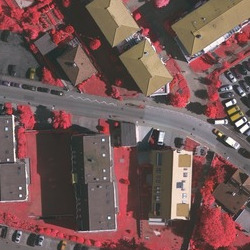}
    \caption{IRRG data}
  \end{subfigure}
%   \begin{subfigure}[t]{0.19\textwidth}
%     \includegraphics[width=\textwidth]{svl}
%     \caption{``SVL''\cite{gerke_use_2015}}
%   \end{subfigure}
%    \begin{subfigure}[t]{0.15\textwidth}
%    \includegraphics[width=\textwidth]{rf}
%    \caption{RF + CRF\cite{quang_efficient_2015}}
%  \end{subfigure}
    \begin{subfigure}[t]{0.15\textwidth}
    \includegraphics[width=\textwidth]{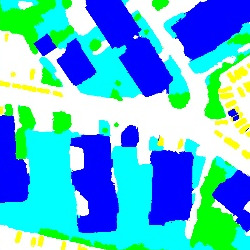}
    \caption{``DLR'' (FCN)\cite{marmanis_semantic_2016}}
  \end{subfigure}
    \begin{subfigure}[t]{0.15\textwidth}
    \includegraphics[width=\textwidth]{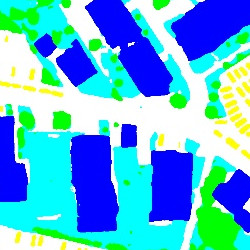}
    \caption{Ours (SegNet)}
  \end{subfigure}
  \caption{Comparison of our method with a FCN on the ISPRS Vaihingen benchmark. Building detection is not impeded by shadows anymore and cars are more finely segmented.\\
  (white: roads, {\color{blue!80!black} blue}: buildings, {\color{cyan!80!black} cyan}: low vegetation, {\color{green!80!black} green}:~trees, {\color{yellow!80!black} yellow}: cars)}
  \label{fig:segnet_qualitative}
\end{figure}

Our best model achieves state-of-the art results on the ISPRS Vaihingen dataset (cf. \cref{tab:leaderboard}) \footnote{\url{http://www2.isprs.org/vaihingen-2d-semantic-labeling-contest.html}}. \cref{fig:segnet_qualitative} illustrates a qualitative comparison between our SegNet-based residual correction and a traditional fully convolutional architecture on an extract of the Vaihingen testing set. The provided metrics are the global pixel-wise overall accuracy (OA) and the F1 score on each class:

\begin{equation}
F1_{i} = 2~\frac{precision_{i} \times recall_{i}}{precision_{i} + recall_{i}}
\end{equation}
\begin{equation}
recall_i = \frac{tp_i}{C_i},~ precision_i = \frac{tp_i}{P_i}~,
\end{equation}

where $tp_i$ the number of true positives for class $i$, $C_i$ the number of pixels belonging to class $i$, and $P_i$ the number of pixels attributed to class $i$ by the model. These metrics are computed using an alternative ground truth in which the borders have been eroded by a 3px radius circle.

\subsection{Analysis}

\begin{figure}[t]
  \captionsetup[subfigure]{singlelinecheck=off,justification=centering}
  \captionsetup[subfigure]{labelformat=empty}
  \centering
  \begin{subfigure}[t]{0.12\textwidth}
    \includegraphics[width=\textwidth]{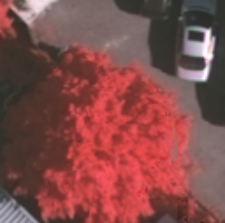}
    \caption{IRRG data}
  \end{subfigure}
  \begin{subfigure}[t]{0.12\textwidth}
    \includegraphics[width=\textwidth]{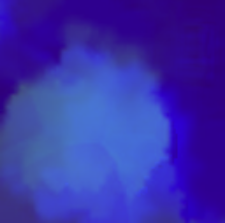}
    \caption{Composite data}
  \end{subfigure}
  \begin{subfigure}[t]{0.12\textwidth}
    \includegraphics[width=\textwidth]{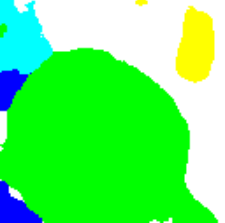}
    \caption{IRRG prediction}
  \end{subfigure}
  \begin{subfigure}[t]{0.12\textwidth}
    \includegraphics[width=\textwidth]{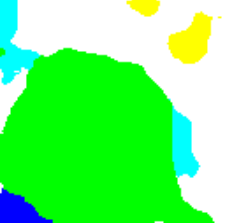}
    \caption{Composite prediction}
  \end{subfigure}
%   \begin{subfigure}[t]{0.13\textwidth}
%     \includegraphics[width=\textwidth]{42_pred_sum.png}
%     \caption{Fusion (average)}
%   \end{subfigure}
  \begin{subfigure}[t]{0.12\textwidth}
    \includegraphics[width=\textwidth]{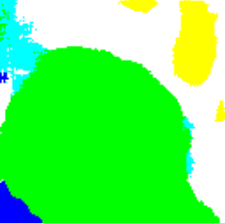}
    \caption{Fusion (network)}
  \end{subfigure} 
  \begin{subfigure}[t]{0.12\textwidth}
    \includegraphics[width=\textwidth]{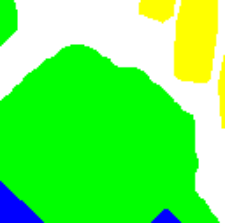}
    \caption{Ground truth}
  \end{subfigure}
  %%%
    \begin{subfigure}[t]{0.12\textwidth}
    \includegraphics[width=\textwidth]{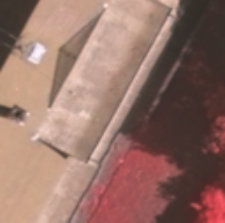}
    \caption{IRRG data}
  \end{subfigure}
  \begin{subfigure}[t]{0.12\textwidth}
    \includegraphics[width=\textwidth]{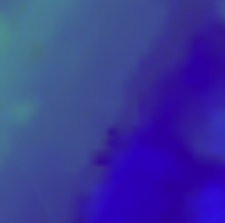}
    \caption{Composite data}
  \end{subfigure}
  \begin{subfigure}[t]{0.12\textwidth}
    \includegraphics[width=\textwidth]{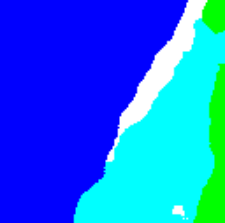}
    \caption{IRRG prediction}
  \end{subfigure}
  \begin{subfigure}[t]{0.12\textwidth}
    \includegraphics[width=\textwidth]{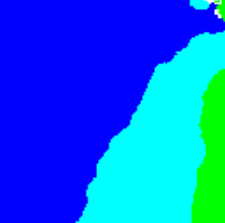}
    \caption{Composite prediction}
  \end{subfigure}
%   \begin{subfigure}[t]{0.13\textwidth}
%     \includegraphics[width=\textwidth]{300_pred_sum}
%     \caption{Fusion (average)}
%   \end{subfigure}
  \begin{subfigure}[t]{0.12\textwidth}
    \includegraphics[width=\textwidth]{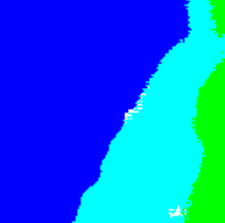}
    \caption{Fusion (network)}
  \end{subfigure} 
  \begin{subfigure}[t]{0.12\textwidth}
    \includegraphics[width=\textwidth]{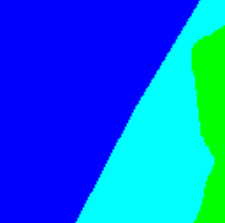}
    \caption{Ground truth}
  \end{subfigure}  
  \caption{Effects of our fusion strategies on selected patches.}
  \label{fig:patches_fusion}
\end{figure}

On the validation set, naive fusion by averaging the maps boosts the OA by 0.4\%, and the residual correction improves it further by an additional 0.4\%. As illustrated in \cref{fig:patches_fusion}, the fusion manages to correct errors in one model by using information from the other source. The residual correction network generates more visually appealing predictions, as it learns which network to favor for each class. For example, the IRRG data is often right when predicting car pixels, therefore the correction network trusts the IRRG prediction about cars more often. However the composite data has the advantage of the DSM to help distinguishing between low vegetation and trees. Thus, the correction network gives more weight to the predictions of the ``composite SegNet'' for these classes. Interestingly, if $m_{avg}$, $m_{corr}$, $s_{avg}$ and $s_{corr}$ denote the respective mean and standard deviation of the activations after averaging and after correction, we see that $m_{avg} \simeq 1.0,~m_{corr} \simeq 0 \text{ and } s_{avg} \simeq 5,~s_{corr} \simeq 2~$. We conclude that the network actually learnt how to apply small corrections to achieve a higher accuracy, which is in phase with both our expectations and theoretical developments~\cite{he_deep_2016}.

This approach obtains state-of-the-art results on the ISPRS Vaihingen 2D Labeling Challenge at 89.8\% \footnote{``ONE\_7'': \url{https://www.itc.nl/external/ISPRS_WGIII4/ISPRSIII_4_Test_results/2D_labeling_vaih/2D_labeling_Vaih_details_ONE_7/index.html}} (cf. \cref{tab:leaderboard}). F1 scores are significantly improved on buildings and vegetation, thanks to the discriminative power of the DSM and NDVI. However, even though the F1 score on cars is competitive, it is lower than expected. This is partly due by poor position and height values of cars in the DSM, making fusion harder for the network. We wish to investigate this issue further, e.g. by incorporating hard-negative mining to help the fusion module learn how to merge very hetereogeneous predictions. Nonetheless, we argue that such our residual correction module can help improve exploit the complementarity of heteregeneous inputs when one has trained an ensemble of classifiers.

\section{Conclusion}

In this work, we presented a residual correction neural network designed to perform prediction-oriented data fusion of heterogeneous sources. On top of parallelized deep fully convolutional networks, the residual correction improved our semantic labeling model using sensor specific information. Especially, our experimental study showed that the residual correction is able to accurately identify which stream to trust for the different classes. We validated the residual correction technique on Earth Observation data, specifically the ISPRS 2D Vaihingen Semantic Labeling challenge, on which we fused IR/R/G, height information and NDVI data and improved the state-of-the-art by~1\%.

Future work involves using residual correction to merge streams coming from networks with different topologies and making the fusion more aware of the early layers, in order to benefit from a mix of low, medium and high level features. We also would like to show that this solution generalizes to other use cases of fusing predictions from several classifiers.


\section*{Acknowledgment}
The Vaihingen dataset was provided by the German Society for Photogrammetry, Remote Sensing and Geoinformation (DGPF) \cite{cramer_dgpf_2010}: \url{http://www.ifp.uni-stuttgart.de/dgpf/DKEP-Allg.html}. Nicolas Audebert's work is supported by the Total-ONERA research project NAOMI.



%\bibliographystyle{IEEEtran}
%\bibliographystyle{IEEEtran}
\bibliographystyle{IEEEtran}
\bibliography{IEEEabrv,JURSE17}

\begin{thebibliography}{10}
\providecommand{\url}[1]{#1}
\csname url@samestyle\endcsname
\providecommand{\newblock}{\relax}
\providecommand{\bibinfo}[2]{#2}
\providecommand{\BIBentrySTDinterwordspacing}{\spaceskip=0pt\relax}
\providecommand{\BIBentryALTinterwordstretchfactor}{4}
\providecommand{\BIBentryALTinterwordspacing}{\spaceskip=\fontdimen2\font plus
\BIBentryALTinterwordstretchfactor\fontdimen3\font minus
  \fontdimen4\font\relax}
\providecommand{\BIBforeignlanguage}[2]{{%
\expandafter\ifx\csname l@#1\endcsname\relax
\typeout{** WARNING: IEEEtran.bst: No hyphenation pattern has been}%
\typeout{** loaded for the language `#1'. Using the pattern for}%
\typeout{** the default language instead.}%
\else
\language=\csname l@#1\endcsname
\fi
#2}}
\providecommand{\BIBdecl}{\relax}
\BIBdecl

\bibitem{everingham_pascal_2014}
M.~Everingham, S.~M.~A. Eslami, L.~Gool, C.~Williams, J.~Winn, and
  A.~Zisserman, ``\BIBforeignlanguage{en}{The {Pascal} {Visual} {Object}
  {Classes} {Challenge}: {A} {Retrospective}},''
  \emph{\BIBforeignlanguage{en}{International Journal of Computer Vision}},
  vol. 111, no.~1, pp. 98--136, Jun. 2014.

\bibitem{rottensteiner_isprs_2012}
F.~Rottensteiner, G.~Sohn, J.~Jung, M.~Gerke, C.~Baillard, S.~Benitez, and
  U.~Breitkopf, ``The {ISPRS} benchmark on urban object classification and 3d
  building reconstruction,'' \emph{ISPRS Ann. Photogramm. Remote Sens. Spat.
  Inf. Sci}, vol.~1, p.~3, 2012.

\bibitem{paisitkriangkrai_effective_2015}
S.~Paisitkriangkrai, J.~Sherrah, P.~Janney, and A.~Van Den~Hengel, ``Effective
  semantic pixel labelling with convolutional networks and {Conditional}
  {Random} {Fields},'' in \emph{Proceedings of the {IEEE} {Conference} on
  {Computer} {Vision} and {Pattern} {Recognition} {Workshops}}, Jun. 2015, pp.
  36--43.

\bibitem{marmanis_semantic_2016}
D.~Marmanis, J.~D. Wegner, S.~Galliani, K.~Schindler, M.~Datcu, and U.~Stilla,
  ``Semantic {Segmentation} of {Aerial} {Images} with an {Ensemble} of
  {CNNs},'' \emph{ISPRS Annals of Photogrammetry, Remote Sensing and Spatial
  Information Sciences}, vol.~3, pp. 473--480, 2016.

\bibitem{marmanis_deep_2016}
D.~Marmanis, M.~Datcu, T.~Esch, and U.~Stilla, ``Deep {Learning} {Earth}
  {Observation} {Classification} {Using} {ImageNet} {Pretrained} {Networks},''
  \emph{IEEE Geoscience and Remote Sensing Letters}, vol.~13, no.~1, pp.
  105--109, Jan. 2016.

\bibitem{ngiam_multimodal_2011}
J.~Ngiam, A.~Khosla, M.~Kim, J.~Nam, H.~Lee, and A.~Ng, ``Multimodal deep
  learning,'' in \emph{Proceedings of the 28th international conference on
  machine learning ({ICML}-11)}, 2011, pp. 689--696.

\bibitem{eitel_multimodal_2015}
A.~Eitel, J.~Springenberg, L.~Spinello, M.~Riedmiller, and W.~Burgard,
  ``Multimodal deep learning for robust {RGB}-{D} object recognition,'' in
  \emph{Proceedings of the {International} {Conference} on {Intelligent}
  {Robots} and {Systems}}.\hskip 1em plus 0.5em minus 0.4em\relax IEEE, 2015,
  pp. 681--687.

\bibitem{lagrange_benchmarking_2015}
A.~Lagrange, B.~Le~Saux, A.~Beaupere, A.~Boulch, A.~Chan-Hon-Tong, S.~Herbin,
  H.~Randrianarivo, and M.~Ferecatu, ``Benchmarking classification of
  earth-observation data: {From} learning explicit features to convolutional
  networks,'' in \emph{{IEEE} {International} {Geosciences} and {Remote}
  {Sensing} {Symposium} ({IGARSS})}, Jul. 2015, pp. 4173--4176.

\bibitem{mou_spatiotemporal_2016}
L.~Mou and X.~Zhu, ``Spatiotemporal {Scene} {Interpretation} of {Space}
  {Videos} via {Deep} {Neural} {Network} and {Tracklet} {Analysis},'' in
  \emph{{IEEE} {International} {Geosciences} and {Remote} {Sensing} {Symposium}
  ({IGARSS})}, Jul. 2016.

\bibitem{he_deep_2016}
K.~He, X.~Zhang, S.~Ren, and J.~Sun, ``Deep {Residual} {Learning} for {Image}
  {Recognition},'' in \emph{Proceedings of the {IEEE} {Conference} on
  {Computer} {Vision} and {Pattern} {Recognition}}, 2016.

\bibitem{long_fully_2015}
J.~Long, E.~Shelhamer, and T.~Darrell, ``Fully {Convolutional} {Networks} for
  {Semantic} {Segmentation},'' in \emph{Proceedings of the {IEEE} {Conference}
  on {Computer} {Vision} and {Pattern} {Recognition}}, 2015, pp. 3431--3440.

\bibitem{badrinarayanan_segnet:_2015}
V.~Badrinarayanan, A.~Kendall, and R.~Cipolla, ``{SegNet}: {A} {Deep}
  {Convolutional} {Encoder}-{Decoder} {Architecture} for {Image}
  {Segmentation},'' \emph{arXiv preprint arXiv:1511.00561}, 2015.

\bibitem{gerke_use_2015}
M.~Gerke, ``Use of the {Stair} {Vision} {Library} within the {ISPRS} 2d
  {Semantic} {Labeling} {Benchmark} ({Vaihingen}),'' International Institute
  for Geo-Information Science and Earth Observation, Tech. Rep., 2015.

\bibitem{quang_efficient_2015}
N.~T. Quang, N.~T. Thuy, D.~V. Sang, and H.~T.~T. Binh, ``An {Efficient}
  {Framework} for {Pixel}-wise {Building} {Segmentation} from {Aerial}
  {Images},'' in \emph{Proceedings of the {Sixth} {International} {Symposium}
  on {Information} and {Communication} {Technology}}.\hskip 1em plus 0.5em
  minus 0.4em\relax ACM, 2015, p.~43.

\bibitem{cramer_dgpf_2010}
M.~Cramer, ``The {DGPF} test on digital aerial camera evaluation – overview
  and test design,'' \emph{Photogrammetrie – Fernerkundung –
  Geoinformation}, vol.~2, pp. 73--82, 2010.

\end{thebibliography}

\end{document}